\documentclass{article}
\usepackage{appendix}
\usepackage{authblk}
\usepackage{amsfonts}
\usepackage{amsmath}
\usepackage{amssymb}
\usepackage{caption}
\usepackage{graphicx,subfig}
\usepackage{float}
\usepackage{algorithm}
\usepackage{algorithmicx}
\usepackage{algpseudocode}
\usepackage{amsmath}
\author[1,2]{Fuzhou Gong \thanks{Corresponding author: fzgong@amt.ac.cn}}
\author[2]{Zigeng Xia \thanks{Corresponding author: xiazigeng14@mails.ucas.edu.cn}}
\affil[1]{Academy of Mathematics and Systems Science, Chinese Academy of Sciences}
\affil[2]{University of Chinese Academy of Sciences}
\title{Generate the corresponding Image from Text
Description using
Modified GAN-CLS Algorithm
}
\begin{document}
\maketitle
\begin{abstract}
Synthesizing images or texts automatically is a useful research area in the artificial intelligence nowadays. Generative adversarial networks (GANs), which are proposed by Goodfellow in 2014, make this task to be done more efficiently by using deep neural networks. We consider generating corresponding images from an input text description using a GAN. In this paper, we analyze the GAN-CLS algorithm, which is a kind of advanced method of GAN proposed by Scott Reed in 2016. First, we find the problem with this algorithm through inference. Then we correct the GAN-CLS algorithm according to the inference by modifying the objective function of the model. Finally, we do the experiments on the Oxford-102 dataset and the CUB dataset. As a result, our modified algorithm can generate images which are more plausible than the GAN-CLS algorithm in some cases. Also, some of the generated images match the input texts better.
\end{abstract}
\section{Introduction}
We focus on generating images from a single-sentence text description in this paper. The Generative adversarial net\cite{ref1} is a widely used generative model in image synthesis. It performs well on many public data sets, the images generated by it seem plausible for human beings. For the original GAN, we have to enter a random vector with a fixed distribution to it and then get the resulting sample. This means that we can not control what kind of samples will the network generates directly because we do not know the correspondence between the random vectors and the result samples. Therefore the conditional GAN (cGAN) \cite{ref2} is proposed. The condition, which may be the class label or the text description, is added to the inputs of both generator and discriminator. As a result, cGAN can generate samples conform with the condition.
\\The GAN-CLS algorithm\cite{ref3} is based on the cGAN. It has one more term in the objective function which contains the mismatched pairs of texts and images. We can infer this algorithm just like the original GAN. But we find that when the objective function in GAN-CLS algorithm achieves its optimum point, the distribution of the generated samples is not the same as the distribution of the data, which is different from the result of original GAN.
Our contribution in this paper is that we modify the objective function of the GAN-CLS algorithm in order to correct it theoretically. We will show the proof of the modified GAN-CLS algorithm and our experimental result on Oxford-102 flower dataset and CUB dataset.
\section{Background}
\subsection{Generative adversarial networks}
Generative adversarial network(GAN) is proposed by Goodfellow in 2014, which is a kind of generative model. It consists of a discriminator network $D$ and a generator network $G$. The input of the generator is a random vector $z$ from a fixed distribution such as normal distribution and the output of it is an image. The input of discriminator is an image , the output is a value in $(0,1)$. The two networks compete during training, the objective function of GAN is:
\begin{align}
\min_{G}\max_{D}V(D,G)=\min_{G}\max_{D}\mathbb{E}_{x{\sim}p_d(x)}[logD(x)]+\mathbb{E}_{z{\sim}p_z(z)}[log(1-D(G(z)))].
\end{align}
In this function, $p_d(x)$ denotes the distribution density function of data samples, $p_z(z)$ denotes the distribution density function of random vector $z$. During the training of GAN, we first fix $G$ and train $D$, then fix $D$ and train $G$. According to\cite{ref1}, when the algorithm converges, the generator can generate samples which obeys the same distribution with the samples from data set.
\\In order to generate samples with restrictions, we can use conditional generative adversarial network(cGAN). cGAN add condition $c$ to both of the discriminator and the generator networks. The condition $c$ can be class label or the text description. The objective function of cGAN is:
\begin{align}
\min_{G}\max_{D}\mathbb{E}_{(x,c){\sim}p_d(x,c)}[logD(x,c)]+\mathbb{E}_{z{\sim}p_z(z),c{\sim}p_d(c)}[log(1-D(G(z,c)),c)].
\end{align}
\subsection{Matching-aware discriminator(GAN-CLS)}
The GAN-CLS algorithm is established base on cGAN and the objective function is modified in order to make the discriminator be matching-aware, which means that the discriminator can judge whether the input text and the image matching. This algorithm is also used by some other GAN based models like StackGAN\cite{ref4}.

The objective function of this algorithm is:
\begin{equation}
\begin{aligned}
\min_{G}\max_{D}&\mathbb{E}_{(x,h){\sim}p_d(x,h)}[logD(x,h)]+\frac{1}{2}\mathbb{E}_{z{\sim}p_z(z),h{\sim}p_d(h)}[log(1-D(G(z,h)),h)]
\\&+\frac{1}{2}\mathbb{E}_{(x,h){\sim}p_{\hat{d}}(x,h)}[log(1-D(x,h))].
\end{aligned}
\end{equation}
In the function, $h$ is the embedding of the text. $p_d(x,h)$ is the distribution density function of the samples from the dataset, in which $x$ and $h$ are matched. $p_{\hat{d}}(x,h)$ is the distribution density function of the samples from dataset consisting of text and mismatched image.
\\The network structure of GAN-CLS algorithm is:
\begin{figure}[H]
  \centering
  \includegraphics[width=0.6\textwidth]{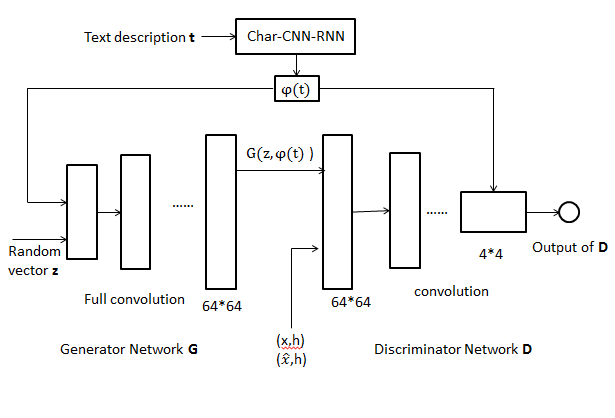}\\
  \caption{Network Structure of GAN-CLS algorithm}\label{structure}
\end{figure}
During training, the text is encoded by a pre-train deep convolutional-recurrent text encoder\cite{ref5}. The discriminator has 3 kinds of inputs: matching pairs of image and text $(x,h)$ from dataset, text and wrong image $(\hat{x},h)$ from dataset, text and corresponding generated image $(G(z,h),h)$.
\section{Method}
\subsection{Problem of GAN-CLS algorithm}
We can infer GAN-CLS algorithm theoretically. Then we have the following theorem:
\paragraph{Theorem 1}
Let the distribution density function of $D(x,h)$ when $(x,h){\sim}p_d(x,h)$ be $f_d(y)$, the distribution density function of $D(x,h)$ when $(x,h){\sim}p_{\hat{d}}(x,h)$ be $f_{\hat{d}}(y)$, the distribution density function of $D(G(z,h),h)$ when\\$z{\sim}p_z(z),h{\sim}p_d(h)$ be $f_g(y)$. Then in the training process of the GAN-CLS algorithm, when the generator is fixed, the form of optimal discriminator is:
\begin{align}
D_{G}^{*}=arg\max_{D}V(D,G)=\frac{f_d(y)}{f_d(y)+\frac{1}{2}(f_{\hat{d}}(y)+f_g(y))}.
\end{align}
The global minimum of $V(D^{*}_{G},G)$ is achieved when the generator $G$ satisfies
\begin{align}
f_g(y)=2f_d(y)-f_{\hat{d}}(y).
\end{align}
\paragraph{Proof} See Appendix A.
\\From this theorem we can see that the global optimum of the objective function is not $f_g(y)=f_d(y)$. This is different from the original GAN. As a result, the generator is not able to generate samples which obey the same distribution with the training data in the GAN-CLS algorithm.
\\But in practice, the GAN-CLS algorithm is able to achieve the goal of synthesizing corresponding image from given text description. We guess the reason is that for the dataset, the distribution $p_d(x)$ and $p_{\hat{d}}(x)$ are similar. Therefore we have $f_g(y)=2f_d(y)-f_{\hat{d}}(y)=f_d(y)$ approximately.
\subsection{Modified GAN-CLS algorithm}
Since the GAN-CLS algorithm has such problem, we propose modified GAN-CLS algorithm to correct it. The method is that we modify the objective function of the algorithm. The definition of the symbols is the same as the last section. Let the distribution density function of $D(x,h)$ when $(x,h){\sim}p_d(x,h)$ be $f_d(y)$, the distribution density function of $D(x,h)$ when $(x,h){\sim}p_{\hat{d}}(x,h)$ be $f_{\hat{d}}(y)$, the distribution density function of $D(G(z,h),h)$ when\\$z{\sim}p_z(z),h{\sim}p_d(h)$ be $f_g(y)$.
\paragraph{Theorem 2}
When we use the following objective function for the discriminator and the generator:
\begin{equation}
\begin{aligned}
V(D,G)&=\frac{1}{2}\{\mathbb{E}_{(x,h){\sim}p_d(x,h)}[logD(x,h)]+\mathbb{E}_{z{\sim}p_z(z),h{\sim}p_d(h)}[log(1-D(G(z,h)),h)]
\\+&\mathbb{E}_{(x,h){\sim}p_{\hat{d}}(x,h)}[log(1-D(x,h))+log(D(x,h))]\},
\end{aligned}
\end{equation}
the form of the optimal discriminator under the fixed generator $G$ is:
\begin{align}
D_{G}^{*}=arg\max_{D}V(D,G)=1-\frac{f_{\hat{d}}(y)+f_g(y)}{2f_{\hat{d}}(y)+f_d(y)+f_g(y)}.
\end{align}
The minimum of the function $V(D_{G}^{*},G)$ is achieved when $G$ satisfies $f_g(y)=f_d(y)$. Then we have
\begin{align}
\min_{G}\max_{D}V(D,G)=-log4.
\end{align}
Which is the same as the original GAN.
\paragraph{Proof}
See Appendix B.
\\The theorem above ensures that the modified GAN-CLS algorithm can do the generation task theoretically. Let $\varphi$ be the encoder for the text descriptions, $G$ be the generator network with parameters $\theta_g$, $D$ be the discriminator network with parameters $\theta_d$, the steps of the modified GAN-CLS algorithm are:
\begin{algorithm}[htb]
\caption{ Modified GAN-CLS algorithm}
\label{alg:Framwork}
\begin{algorithmic}[1]
  \Require
    minibatch size $m$;
    learning rate $\epsilon$;
    number of iterations $N$;
    dataset $X$
  \For{$i$ in 1 to $N$}
     \State extract $m$ samples $\{(x^{(1)}, t^{(1)}), (x^{(2)}, t^{(2)}),..., (x^{(m)} ,t^{(m)})\}$ from one class of the dataset $X$, where $x^{(i)}$ is the image and $t^{(i)}$ is the corresponding text description.
     \State extract $m$ images $\{\hat{x}^{(1)}, \hat{x}^{(2)},..., \hat{x}^{(m)}\}$ from another class in $X$.
     \State encode the text descriptions: $h^{(i)}=\varphi(t^{(i)}),i=1,...,m$.
     \State extract $m$ random vectors $\{z^{(1)}, z^{(2)},..., z^{(m)}\}$ from the distribution $p_z(z)$.
     \State generate images $\tilde{x}^{(i)}=G(z^{(i)},h^{(i)}),i=1,...,m$.
     \State calculate $L_D=-\frac{1}{m}\sum_{i=1}^{m}\frac{1}{2}[log(D(x^{(i)},h^{(i)}))+log(1-D(\tilde{x}^{(i)},h^{(i)}))+log(D(\hat{x}^{(i)},h^{(i)}))+log(1-D(\hat{x}^{(i)},h^{(i)}))]$.
     \State update the parameters of the discriminator: $\theta_d\leftarrow\theta_d-\epsilon{\nabla}_{\theta_d}L_D(\theta_d)$.
     \State calculate $L_G=\frac{1}{m}\sum_{i=1}^{m}\frac{1}{2}log(1-D(\tilde{x}^{(i)},h^{(i)})))$.
     \State update the parameters of the generator: $\theta_g\leftarrow\theta_g-\epsilon{\nabla}_{\theta_g}L_G(\theta_g)$.
  \EndFor
\end{algorithmic}
\end{algorithm}
\section{Experiments}
\subsection{Datasets}
We do the experiments on the Oxford-102 flower dataset and the CUB dataset with GAN-CLS algorithm and modified GAN-CLS algorithm to compare them. For the Oxford-102 dataset, it has 102 classes, which contains 82 training classes and 20 test classes. For the CUB dataset, it has 200 classes, which contains 150 train classes and 50 test classes. Each of the images in the two datasets has 10 corresponding text descriptions. We use the same network structure as well as parameters for both of the datasets. For the Oxford-102 dataset, we train the model for 100 epoches, for the CUB dataset, we train the model for 600 epoches.
\subsection{Other details}
\subsubsection{Sampling method}
We use mini-batches to train the network, the batch size in the experiment is $64$. One mini-batch consists of $64$ three element sets: \{image $x_1$, corresponding text description $t_1$, another image $x_2$\}. Every time we use a random permutation on the training classes, then we choose the first class and the second class. In the first class, we pick image $x_1$ randomly and in the second class we pick image $x_2$ randomly. Then pick one of the text descriptions of image $x_1$ as $t_1$.
\subsubsection{Structure and parameters}
For the network structure, we use DCGAN\cite{ref6}. The size of the generated image is $64*64*3$. We use a pre-trained char-CNN-RNN network to encode the texts. We also use the GAN-INT algorithm proposed by Scott Reed\cite{ref3}. This algorithm calculates the interpolations of the text embeddings pairs and add them into the objective function of the generator:
\begin{align}
\mathbb{E}_{h_1,h_2{\sim}p_d(h)}[log(1-D(G(z,{\alpha}h_1+(1-{\alpha})h_2)))].
\end{align}
There are no corresponding images or texts for the interpolated text embeddings, but the discriminator can tell whether the input image and the text embedding match when we use the modified GAN-CLS algorithm to train it. So doing the text interpolation will enlarge the dataset. We find that the GAN-INT algorithm performs well in the experiments, so we use this algorithm.
Adam algorithm\cite{ref7} is used to optimize the parameters. Learning rate is set to be $0.0002$ and the momentum is $0.5$. The number of filters in the first layer of the discriminator and the generator is $128$. Batch normalization\cite{ref8} is used to make the training more stable and faster.
\subsection{Results}
We enumerate some of the results in our experiment. The two algorithms use the same parameters.
\subsubsection{Training set}
\begin{figure}[H]
  \centering
  \includegraphics[width=1.0\textwidth]{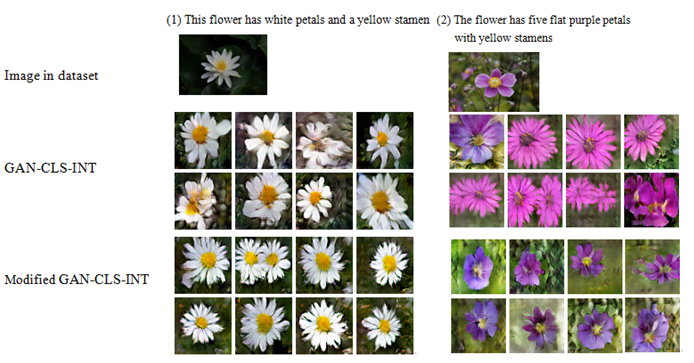}\\
  \caption{Oxford-102 training set result 1}
  \label{train1}
\end{figure}

\begin{figure}[H]
  \centering
  \includegraphics[width=1.0\textwidth]{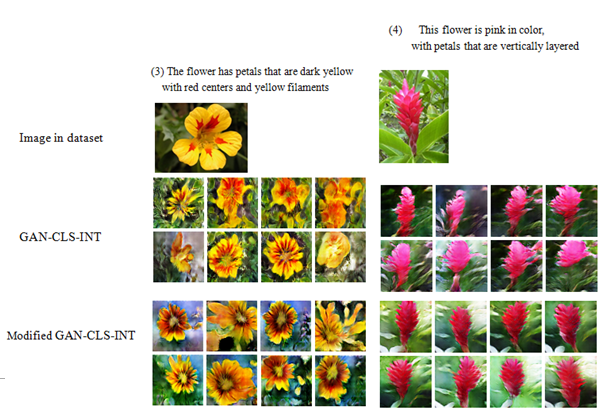}\\
  \caption{Oxford-102 training set result 2}
  \label{train2}
\end{figure}

\begin{figure}[H]
  \centering
  \includegraphics[width=0.6\textwidth]{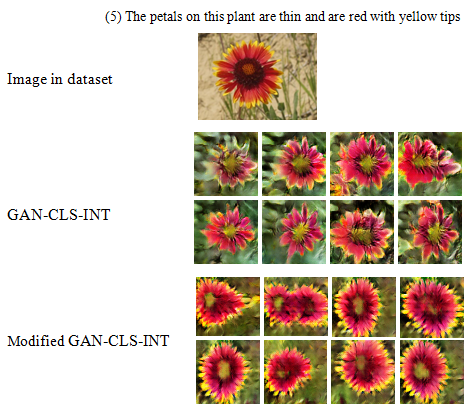}\\
  \caption{Oxford-102 training set result 3}
  \label{train3}
\end{figure}

\begin{figure}[H]
  \centering
  \includegraphics[width=1.0\textwidth]{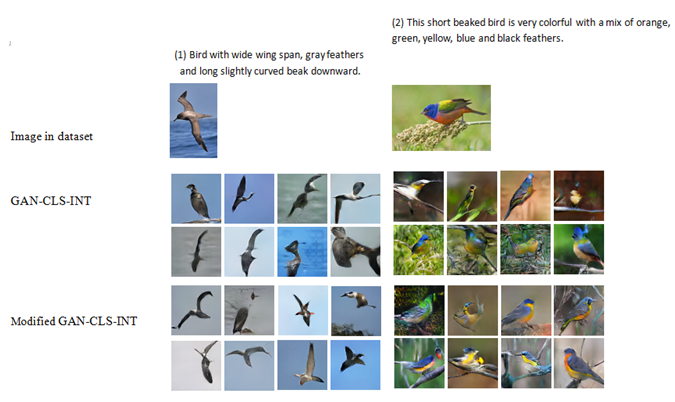}\\
  \caption{CUB training set result 1}
  \label{btrain1}
\end{figure}

\begin{figure}[H]
  \centering
  \includegraphics[width=1.0\textwidth]{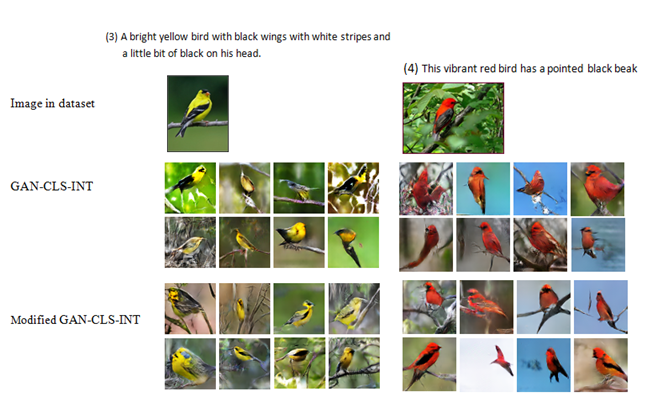}\\
  \caption{CUB training set result 2}
  \label{btrain2}
\end{figure}
For the training set of Oxford-102, In figure 2, we can see that in the result (1), the modified GAN-CLS algorithm generates more plausible flowers. In the result (2), the text contains a detail which is the number of the petals. The images generated by modified algorithm match the text description better. In figure 3, for the result (3), both of the algorithms generate plausible flowers. In the result (4), both of the algorithms generate flowers which are close to the image in the dataset. As for figure 4, the shape of the flower generated by the modified algorithm is better.
\\For the training set of the CUB dataset, we can see in figure 5, In (1), both of the algorithms generate plausible bird shapes, but some of the details are missed. For example, the beak of the bird. In (2), the colors of the birds in our modified algorithm are better. For figure 6, in the result (3), the shapes of the birds in the modified algorithm are better. In (4), the results of the two algorithms are similar, but some of the birds are shapeless.
\subsubsection{Test set}

\begin{figure}[H]
  \centering
  \includegraphics[width=1.0\textwidth]{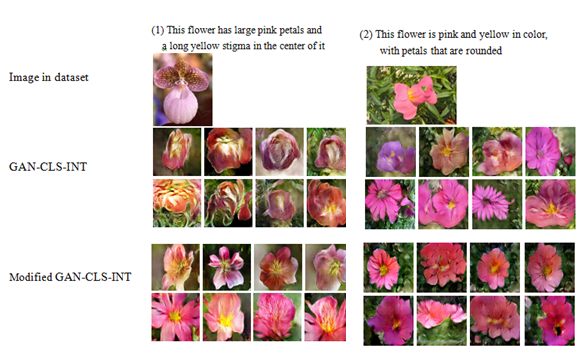}\\
  \caption{Oxford-102 test set result 1}
  \label{test1}
\end{figure}

\begin{figure}[H]
  \centering
  \includegraphics[width=1.0\textwidth]{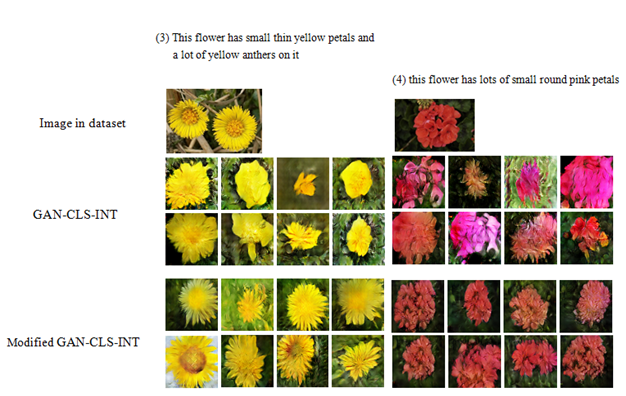}\\
  \caption{Oxford-102 test set result 2}
  \label{test2}
\end{figure}

\begin{figure}[H]
  \centering
  \includegraphics[width=1.0\textwidth]{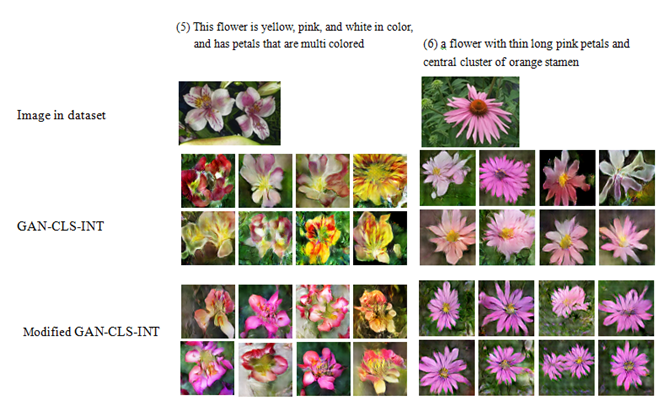}\\
  \caption{Oxford-102 test set result 3}
  \label{test3}
\end{figure}

\begin{figure}[H]
  \centering
  \includegraphics[width=1.0\textwidth]{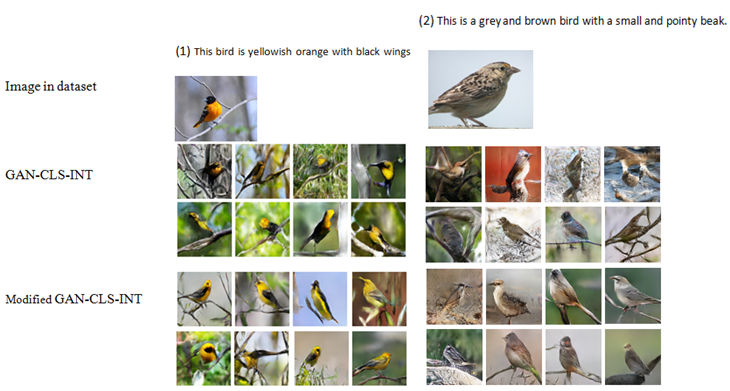}\\
  \caption{CUB test set result 1}
  \label{btest1}
\end{figure}

\begin{figure}[H]
  \centering
  \includegraphics[width=1.0\textwidth]{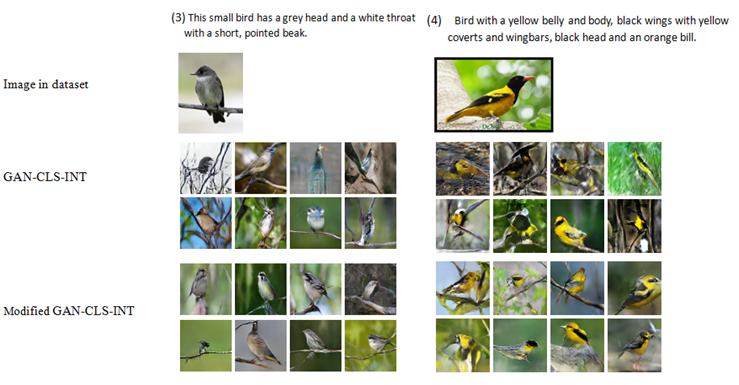}\\
  \caption{CUB test set result 2}
  \label{btest2}
\end{figure}
For the test set, the results are relatively poor in some cases. In the Oxford-102 dataset, we can see that in the result (1) in figure 7, the modified algorithm is better. In (2), the modified algorithm catches the detail "round" while the GAN-CLS algorithm does not. For figure 8, the modified algorithm generates yellow thin petals in the result (3) which match the text better. In (4), both of the algorithms generate images which match the text, but the petals are mussy in the original GAN-CLS algorithm. In (5), the modified algorithm performs better. In (6), the modified algorithm generates more plausible flowers but the original GAN-CLS algorithm can give more diversiform results.
\\In the results of CUB dataset, in (1) of figure 10, the images in the modified algorithm are better and embody the color of the wings. In (2), the images in the modified algorithm are better, which embody the shape of the beak and the color of the bird. For (3) in figure 11, in some results of the modified algorithm, the details like "gray head" and "white throat" are reflected better. In (4), the shapes of the birds are not fine but the modified algorithm is slightly better.
\\According to all the results, both of the algorithms can generate images match the text descriptions in the two datasets we use in the experiment. In some situations, our modified algorithm can provide better results. After training, our model has the generalization ability to synthesise corresponding images from text descriptions which are never seen before.
\subsubsection{Other results}
There are also some results where neither of the GAN-CLS algorithm nor our modified algorithm performs well.
\begin{figure}[H]
  \centering
  \includegraphics[width=0.6\textwidth]{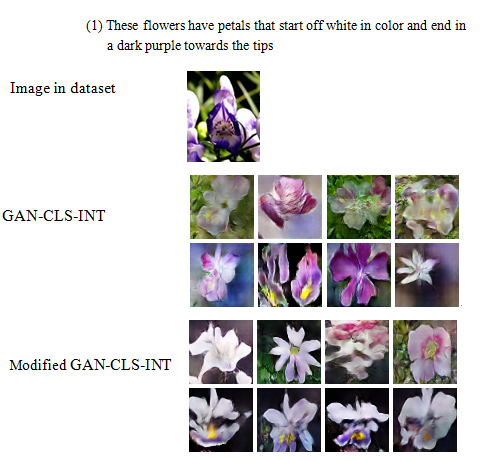}\\
  \caption{An example of bad result}
  \label{bad}
\end{figure}
The text descriptions in these cases are slightly complex and contain more details (like the position of the different colors in Figure 12). We infer that the capacity of our model is not enough to deal with them, which causes some of the results to be poor. Also, the capacity of the datasets is limited, some details may not be contained enough times for the model to learn.
\\For the guess in the last paragraph of section 3.1, we do the following experiment: For the image in the mismatched pairs, we segment it into $16$ pieces, then exchange some of them. After doing this, the distribution $p_d$ and $p_{\hat{d}}$ will not be similar any more. Then we train the model using two algorithms. Our manipulation of the image is shown in figure 13 and we use the same way to change the order of the pieces for all of the images in distribution $p_{\hat{d}}$.
\begin{figure}[H]
  \centering
  \includegraphics[width=0.8\textwidth]{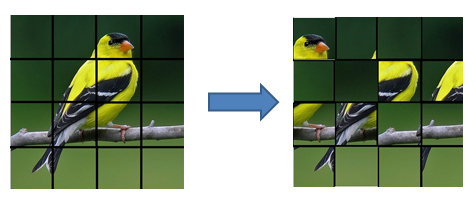}\\
  \caption{An example of manipulation on the mismatched image}
  \label{example}
\end{figure}
Some of the results we get in this experiment are:
\begin{figure}[H]
  \centering
  \includegraphics[width=1.0\textwidth]{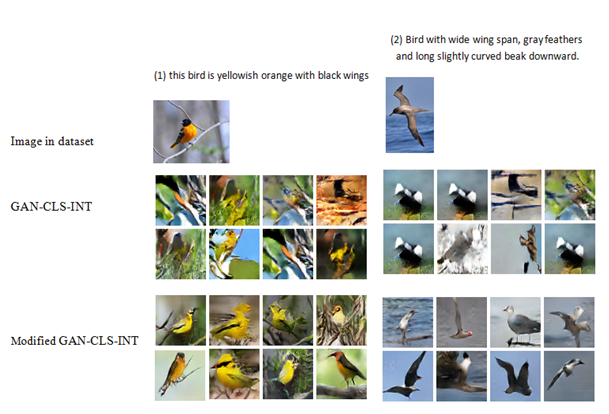}\\
  \caption{Results for the turbulent mismatched images}
  \label{wrong1}
\end{figure}
\begin{figure}[H]
  \centering
  \includegraphics[width=0.6\textwidth]{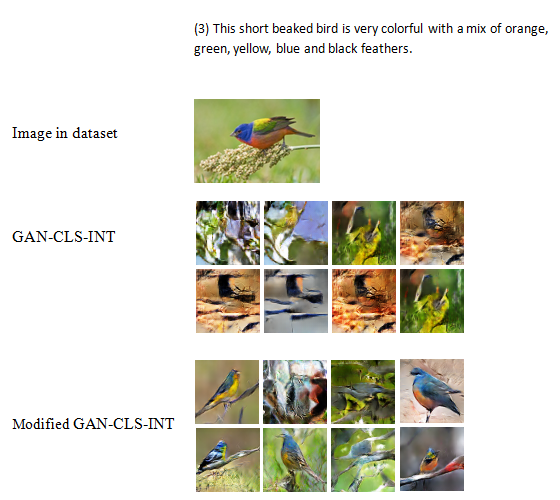}\\
  \caption{Results for the turbulent mismatched images}
  \label{wrong2}
\end{figure}
In these results, the modified GAN-CLS algorithm can still generate images as usual. The results are similar to what we get on the original dataset. However, the original GAN-CLS algorithm can not generate birds anymore. This is consistent with the theory, in the dataset where the distribution $p_d$ and $p_{\hat{d}}$ are not similar, our modified algorithm is still correct. But the generated samples of original algorithm do not obey the same distribution with the data.
\section{Conclusion}
In this paper, we point out the problem of the GAN-CLS algorithm and propose the modified algorithm. The theoretical analysis ensures the validity of the modified algorithm. In the mean time, the experiment shows that our algorithm can also generate the corresponding image according to given text in the two datasets. However, there are still some defects in our algorithm:
\\(1) In some cases, the results of generating are not plausible. The flower or the bird in the image is shapeless, without clearly defined boundary.
\\(2) The algorithm is sensitive to the hyperparameters and the initialization of the parameters. In the experiment, we find that the same algorithm may perform different among several times.
\\(3) The postures of the generated flowers and birds are short of variety, the results seem similar for one fixed text description.
\\In future work, we may try to find the methods to solve these problems.

\begin{appendices}
\section{Proof of Theorem 1}
Firstly, when we fix $G$ and train $D$, we consider:
\begin{equation}
\begin{aligned}
V(D,G)=&\mathbb{E}_{(x,h){\sim}p_d(x,h)}[logD(x,h)]+\frac{1}{2}\mathbb{E}_{z{\sim}p_z(z),h{\sim}p_d(h)}[log(1-D(G(z,h)),h)]
\\&+\frac{1}{2}\mathbb{E}_{(x,h){\sim}p_{\hat{d}}(x,h)}[log(1-D(x,h))].
\end{aligned}
\end{equation}
We assume function $f_d(y)$, $f_g(y)$ and $f_{\hat{d}}(y)$ have the same support set $(0,1)$. Then
\begin{align}
V(D,G)=&\int_{0}^{1}f_d(y)log(y)dy+\frac{1}{2}\int_{0}^{1}(f_{\hat{d}}(y)+f_g(y))log(1-y)dy
\\=&\int_{0}^{1}f_d(y)log(y)+\frac{1}{2}(f_{\hat{d}}(y)+f_g(y))log(1-y)dy.
\end{align}
Since the maximum of function $alog(y)+blog(1-y)$ is achieved when $y=\frac{a}{a+b}$ with respect to $y\in(0,1)$, we have the inequality:
\begin{equation}
\begin{aligned}
V(G,D)&\le\int_{0}^{1}log(\frac{f_d(y)}{f_d(y)+\frac{1}{2}(f_{\hat{d}}(y)+f_g(y))})f_d(y)
\\&+\frac{1}{2}log(\frac{\frac{1}{2}(f_{\hat{d}}(y)+f_g(y))}{f_d(y)+\frac{1}{2}(f_{\hat{d}}(y)+f_g(y))})(f_{\hat{d}}(y)+f_g(y))dy.
\end{aligned}
\end{equation}
When the equality is established, the optimal discriminator is:
\begin{align}
D_{G}^{*}=\frac{f_d(y)}{f_d(y)+\frac{1}{2}(f_{\hat{d}}(y)+f_g(y))}.
\end{align}
Secondly, we fix the discriminator and train the generator.
\begin{equation}
\begin{aligned}
V(G,D)&{\le}V(D^{*}_{G},G)=\int_{0}^{1}log(\frac{f_d(y)}{\frac{1}{2}(f_d(y)+\frac{1}{2}(f_{\hat{d}}(y)+f_g(y)))})f_d(y)
\\&+\frac{1}{2}log(\frac{\frac{1}{2}(f_{\hat{d}}(y)+f_g(y))}{\frac{1}{2}(f_d(y)+\frac{1}{2}(f_{\hat{d}}(y)+f_g(y)))})(f_{\hat{d}}(y)+f_g(y))dy-log4.
\end{aligned}
\end{equation}
Then we have:
\begin{equation}
\begin{aligned}
V(D^{*}_{G},G)&=KL(f_d(y)||\frac{1}{2}(f_d(y)+\frac{1}{2}(f_{\hat{d}}(y)+f_g(y))))
\\&+KL(\frac{1}{2}(f_{\hat{d}}(y)+f_g(y))||\frac{1}{2}(f_d(y)+\frac{1}{2}(f_{\hat{d}}(y)+f_g(y))))-log4
\end{aligned}
\end{equation}
\begin{align}
&=2JSD(f_d(y)||\frac{1}{2}(f_{\hat{d}}(y)+f_g(y)))-log4.
\end{align}
Where $KL(P||Q)$ denotes the Kullback-Leibler divergence, $JSD(P||Q)$ denotes the Jensen-Shannon divergence. $P$ and $Q$ are distribution density functions. Since $JSD(P||Q)\ge0$ and $JSD(P||Q)=0$ if and only if $P=Q$, we have:
\begin{align}
V(D^{*}_{G},G)\ge-log4.
\end{align}
Function $V(D^{*}_{G},G)$ achieves its minimum $-log4$ if and only if $G$ satisfies that $f_d(y)=\frac{1}{2}(f_{\hat{d}}(y)+f_g(y))$, which is equivalent to $f_g(y)=2f_d(y)-f_{\hat{d}}(y)$. This finishes the proof of theorem 1.
\section{Proof of Theorem 2}
First we have
\begin{align}
V(D,G)=\int_{0}^{1}\frac{1}{2}(f_d(y)+f_{\hat{d}}(y))log(y)dy+\int_{0}^{1}\frac{1}{2}(f_g(y)+f_{\hat{d}}(y))log(1-y)dy,
\end{align}
then the same method as the proof for theorem 1 will give us the form of the optimal discriminator:
\begin{align}
D_{G}^{*}=arg\max_{D}V(D,G)=&\frac{\frac{1}{2}(f_d(y)+f_{\hat{d}}(y))}{\frac{1}{2}(f_d(y)+f_{\hat{d}}(y))+\frac{1}{2}(f_g(y)+f_{\hat{d}}(y))}
\\&=1-\frac{f_{\hat{d}}(y)+f_g(y)}{2f_{\hat{d}}(y)+f_d(y)+f_g(y)}.
\end{align}
For the optimal discriminator, the objective function is:
\begin{equation}
\begin{aligned}
V(D_{G}^{*},G)=&\int_{0}^{1}\frac{1}{2}(f_d(y)+f_{\hat{d}}(y))log\frac{\frac{1}{2}(f_d(y)+f_{\hat{d}}(y))}{\frac{1}{2}(f_d(y)+f_{\hat{d}}(y))+\frac{1}{2}(f_g(y)+f_{\hat{d}}(y))}dy
\\+&\int_{0}^{1}\frac{1}{2}(f_g(y)+f_{\hat{d}}(y))log\frac{\frac{1}{2}(f_g(y)+f_{\hat{d}}(y))}{\frac{1}{2}(f_d(y)+f_{\hat{d}}(y))+\frac{1}{2}(f_g(y)+f_{\hat{d}}(y))}dy
\end{aligned}
\end{equation}
\begin{equation}
\begin{aligned}
\\=&-log4+\int_{0}^{1}\frac{1}{2}(f_d(y)+f_{\hat{d}}(y))log\frac{\frac{1}{2}(f_d(y)+f_{\hat{d}}(y))}{\frac{1}{2}(\frac{1}{2}(f_d(y)+f_{\hat{d}}(y))+\frac{1}{2}(f_g(y)+f_{\hat{d}}(y)))}dy
\\+&\int_{0}^{1}\frac{1}{2}(f_g(y)+f_{\hat{d}}(y))log\frac{\frac{1}{2}(f_g(y)+f_{\hat{d}}(y))}{\frac{1}{2}(\frac{1}{2}(f_d(y)+f_{\hat{d}}(y))+\frac{1}{2}(f_g(y)+f_{\hat{d}}(y)))}dy
\end{aligned}
\end{equation}
\begin{equation}
\begin{aligned}
=&-log4+KL(\frac{1}{2}(f_d(y)+f_{\hat{d}}(y))||\frac{1}{2}(\frac{1}{2}(f_d(y)+f_{\hat{d}}(y))+\frac{1}{2}(f_g(y)+f_{\hat{d}}(y))))
\\+&KL(\frac{1}{2}(f_g(y)+f_{\hat{d}}(y))||\frac{1}{2}(\frac{1}{2}(f_d(y)+f_{\hat{d}}(y))+\frac{1}{2}(f_g(y)+f_{\hat{d}}(y))))
\end{aligned}
\end{equation}
\begin{align}
=-log4+2JSD(\frac{1}{2}(f_d(y)+f_{\hat{d}}(y))||\frac{1}{2}(f_g(y)+f_{\hat{d}}(y))).
\end{align}
The minimum of the JS-divergence in (25) is achieved if and only if $\frac{1}{2}(f_d(y)+f_{\hat{d}}(y))=\frac{1}{2}(f_g(y)+f_{\hat{d}}(y))$, this is equivalent to $f_g(y)=f_d(y)$. The generator in the modified GAN-CLS algorithm can generate samples which obeys the same distribution with the sample from dataset. The optimum of the objective function is:
\begin{align}
\min_{G}\max_{D}V(D,G)=-log4.
\end{align}
\end{appendices}
\end{document}